\documentclass{article}
\usepackage[a4paper, total={7in, 10in}]{geometry}

\usepackage[mathscr]{eucal}
\usepackage{amsthm}
\usepackage{graphicx}
\usepackage{mathtools}
\usepackage[hyphens]{url} 
\usepackage[super]{nth}
\usepackage{authblk}

\usepackage{multirow}
\usepackage{xcolor,colortbl}
\usepackage{amsmath}
\usepackage{amsfonts}

\usepackage{algcompatible}
\usepackage{algorithm}
\usepackage{xspace}

\usepackage[norule, flushmargin]{footmisc}

\usepackage{framed}
\usepackage{booktabs}
\usepackage{hyperref,microtype,subcaption}
\usepackage[onehalfspacing]{setspace}
\usepackage{dcolumn}
\usepackage[normalem]{ulem} 

\usepackage{enumitem}

\newcommand{\cf}{\emph{cf.}~}

\newcommand{\etal}{\emph{et al.}~}

\title{Instructing Text-to-Image Generation Models on Fairness}

\date{}

\author[1,2,*]{Felix Friedrich}
\author[1,3]{Manuel Brack}
\author[1]{Lukas Struppek}
\author[1]{Dominik Hintersdorf}
\author[1,2,3,4]{Patrick Schramowski}
\author[5]{Sasha Luccioni}
\author[1,2,4,6]{Kristian Kersting}

\affil[1]{Technical University of Darmstadt, Artificial Intelligence and Machine Learning Lab, Germany}
\affil[2]{Hessian Center for Artificial Intelligence (hessian.AI), Germany}
\affil[3]{German Center for Artificial Intelligence (DFKI), Germany}
\affil[4]{LAION, online}
\affil[5]{Huggingface, Canada}
\affil[6]{Technical University of Darmstadt, Centre for Cognitive Science, Germany}
\affil[*]{Corresponding author: Felix Friedrich (friedrich@cs.tu-darmstadt.de)}

\begin{document}

\maketitle

\begin{abstract}
\noindent Generative AI models have recently achieved astonishing results in quality and are consequently employed in a fast-growing number of applications. However, since they are highly data-driven, relying on billion-sized datasets randomly scraped from the internet, they also suffer from degenerated and biased human behavior, as we demonstrate. In fact, they may even reinforce such biases. To not only uncover but also combat these undesired effects, we present a novel strategy, called \textsc{Fair Diffusion}, to attenuate biases after the deployment of generative text-to-image models. 
Specifically, we demonstrate shifting a bias, based on human instructions, in any direction yielding arbitrary proportions for, e.g., identity groups. As our empirical evaluation demonstrates, this introduced control enables instructing generative image models on fairness, requiring no data filtering nor additional training.
\end{abstract}

\noindent Artificial intelligence (AI) has become an integral part of our lives. However, the deployment of AI systems has sparked a debate on important ethical concerns, especially around fairness. There is a growing concern that AI systems perpetuate and even amplify existing biases, leading to unfair outcomes.
One key area where fairness is critical is text-to-image (or image-to-image) synthesis \cite{Rombach_2022_CVPR,yu22parti,ramesh2022hierarchical,nichol2022glide,saharia2022photorealistic}, which may revolutionize a range of applications, including marketing and social media. Diffusion Models (DM) like Stable Diffusion (SD) \cite{Rombach_2022_CVPR} are currently a widely used variant of image synthesis models, which generate realistic and high-quality images based on text input.

However, despite these successes and similar to generative language models, they inherently suffer from biased and unfair behavior. In this regard, our work is twofold. For one, we audit generative models for present biases and at the same time evaluate their mitigation. Attempts to address these issues comprise data pre-processing \cite{schramowski2022can, nichol2022glide} and in-process approaches \cite{Li_2022_ECCV,berg-etal-2022-prompt}.
Inspired by advances in instructing AI systems based on human feedback \cite{Ouyang2022TrainingLM, friedrich2022XIL_typo}, we explore the instruction of text-to-image models on fairness in this work. 
For the audit, we evaluate biases in the publicly available text-to-image model Stable Diffusion, its underlying large-scale dataset as well as its pre-trained modules. Therefore, we created a subset of LAION-5B \cite{schuhmann2022laion} containing over 1.8M images depicting over 150 occupations to approximate the present gender occupation bias. Furthermore, we identify potential strategies for addressing and mitigating previously found biases, particularly in a model's generated output. Most importantly, we propose a novel and advanced strategy, \textsc{Fair Diffusion}, for bias mitigation at the deployment stage of generative models. It utilizes a (textual) interface for instructing the model, which we envision as an essential component for designing and implementing fair DMs.

\textsc{Fair Diffusion} builds on biased concepts captured in a model from training and steers them in a given direction to increase fairness at inference. A user is in control and guides the model by instructing on fairness.
For the first time, \textsc{Fair Diffusion} offers a practical approach to fairness in DMs. This way, it is possible to realize different notions of fairness, e.g., outcome impartiality, in a single framework, easily accessible to individuals. 

By addressing these fairness issues, we pave the way for DMs to be used in a way that is more ethical, fair, and beneficial to members of society. More importantly, with our strategy, the user regains some control over the model output, which has previously been ceded to a small number of entities with large computational resources.

To summarize, we make the following contributions, by
\begin{enumerate}[label=(\roman*)]
\item  
inspecting and measuring the sources of (gender occupation) biases using the example of Stable Diffusion, 
\item proposing and evaluating a novel strategy, \textsc{Fair Diffusion}, to overcome and mitigate unfair model outcomes,
\item discussing future pathways for fair diffusion models, specifically how they can be integrated into societies to directly promote fairness with a user in control.
\end{enumerate}
We provide data and code to reproduce our experiments.\footnote{Our code is publicly available at \href{https://github.com/ml-research/Fair-Diffusion}{https://github.com/ml-research/Fair-Diffusion}}

The paper is organized as follows. We start by introducing \textsc{Fair Diffusion}, which enables us to mitigate biases in diffusion models, including the applied definition of fairness. Next, we examine the components of Stable Diffusion for biases and demonstrate their mitigation using \textsc{Fair Diffusion} on the example of gender occupation bias.
Before concluding, we extensively discuss our results and highlight a focal shift in achieving fairness through interaction with biased models after deployment. Finally, we provide a detailed background and the underlying methodology of \textsc{Fair Diffusion} next to presenting methods to inspect biases in each component of Stable Diffusion.


\textbf{Disclaimer:} This paper depicts images of different kinds of biases and stereotypes that some readers may find offensive. 
We emphasize that the goal of this work is to investigate and eventually mitigate these biases, which are already present in generative models.
We do not intend to discriminate against identity groups or cultures in any way.

\section*{Fair Diffusion}
\label{sec:large_pretrained_models}
In the following, we present our novel strategy, \textsc{Fair Diffusion}, for approaching fairness in text-to-image generation models. Therefore, we propose to instruct, as motivated before, text-to-image DMs on fairness with textual guidance. Furthermore, we elaborate on fairness definitions in the scope of investigating DMs. In the Methods section, we further present in more detail the textual interface for DMs and the underlying technique to guide their image generation.

\begin{figure}[t]
    \centering
    \includegraphics[width=0.9\linewidth]{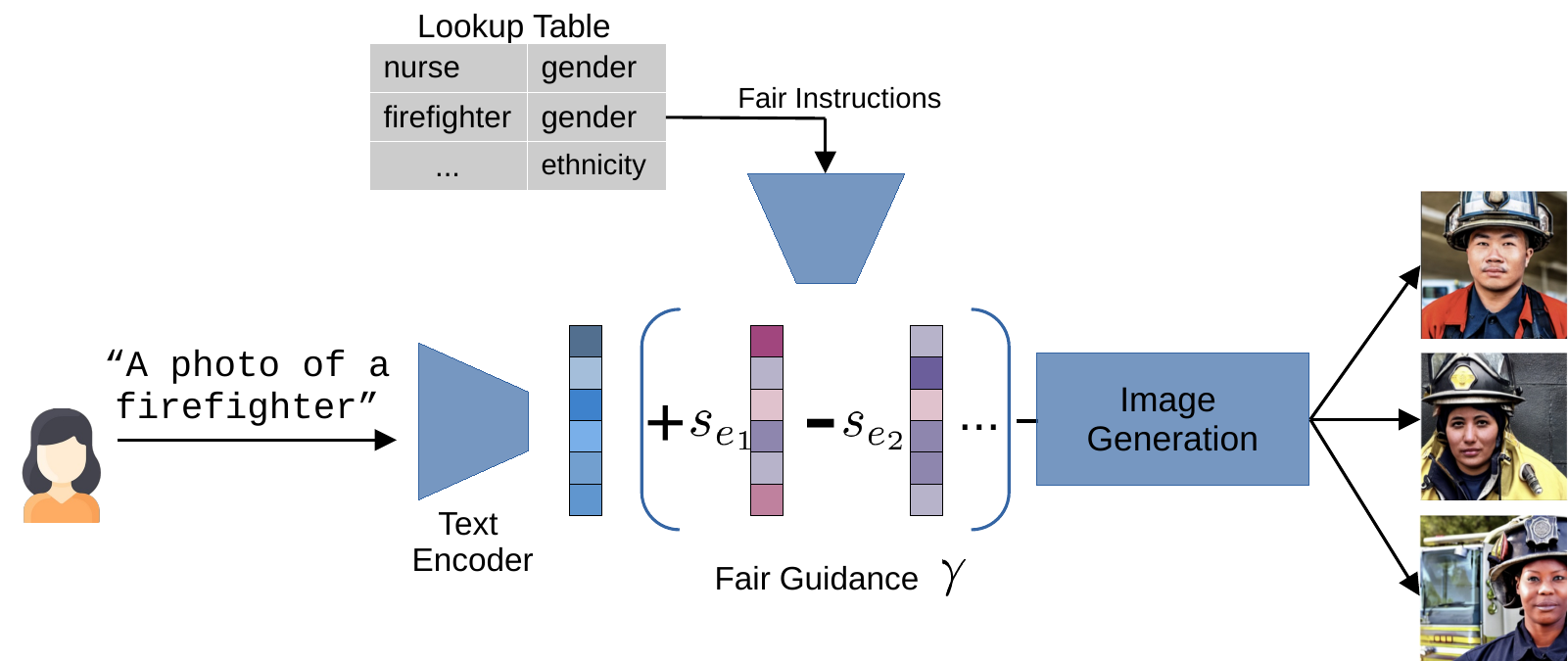}
    \caption{\textsc{Fair Diffusion} deployment. A user inserts a prompt to generate an image. With the help of fair guidance, the image generation is steered toward a fairer outcome. Here, the fair instructions are realized with a lookup table ---the biased concept is recognized, and guidance $\gamma$ is applied. 
    Like in Eq.~\ref{eqn:fairguidance}, the fair instructions, $e_i$, are transformed into vectors, $c_{e_i}$, by the text encoder and can be scaled by $s_{e_i}$ to perform fair guidance. Here two editing prompts (purple-colored vectors) are illustrated, but different numbers are possible too. The lookup table can be set up by any user. (Best viewed in color)}
    \label{fig:deployment}
\end{figure}

\subsubsection*{Instructing Text-to-image Models on Fairness}
\label{sec:instruction}
We propose \textsc{Fair Diffusion} to instruct a pre-trained model on fairness during the deployment stage (\cf Fig.~\ref{fig:deployment}).
In line with this goal, previous work has proposed approaches to alter the image generation process  \cite{hertz2022prompt,liu2022Compositional,brack2023Sega}. \textsc{Fair Diffusion} can facilitate these various approaches as instructing interface. Here, we evaluate \textsc{Fair Diffusion} with Semantic Guidance (\textsc{Sega} \cite{brack2023Sega}) since it enables highly flexible instructions through text (more details in Methods).
With this tool, \textsc{Fair Diffusion} can be applied on a DM as illustrated in Fig.~\ref{fig:deployment}. Based on detected biases---e.g., stored in a lookup table as instructions---the model is guided to reduce bias in its outcomes, eventually moving toward fairer and user-aligned behavior.

Intuitively and on a high level, \textsc{Fair Diffusion} extends classifier-free guidance \cite{ho2022classifier} with an additional fair guidance term $\gamma$ (\cf Eq.~\ref{eqn:fairguidance}) to generate an image $x$:
\begin{equation}
x = \eta(p,\gamma(e,s_e))
\end{equation}
This way, the image generation, $\eta$, is a function of the text input prompt $p$ (e.g., ``A photo of a firefighter'') with fair guidance, $\gamma$, during the generation. In turn, $\gamma$ depends on additional textual descriptions of attribute expressions $e_i$, scaled by $s_{e_i}$ with guidance direction. As a result, the image generation is guided towards the input prompt $p$ and fairness instructions $e_i$ at the same time.
In order to realize the different expressions of an attribute with \textsc{Fair Diffusion}, we control the guidance direction by randomly sampling from a desired probability distribution $P$. That means each $e_i$ is either increased or decreased depending on the expression that should be promoted/ suppressed. In Fig.~\ref{fig:deployment}, we illustrate a binary case, where concept $e_1$ is promoted ($+$) and $e_2$ suppressed ($-$) during the image generation. Their direction can be changed based on $P$.
%
\textsc{Fair Diffusion} can easily facilitate arbitrary distributions by adjusting the guidance probability accordingly. This level of flexibility allows the implementation of different definitions of fairness using the same underlying model. For the evaluation, we again verify the resulting attribute proportion with a classifier. Ideally, the user-defined and measured proportions match.

\begin{figure}
  \centering
  \setlength{\lineskip}{4pt}
  \includegraphics[width=0.92\textwidth]{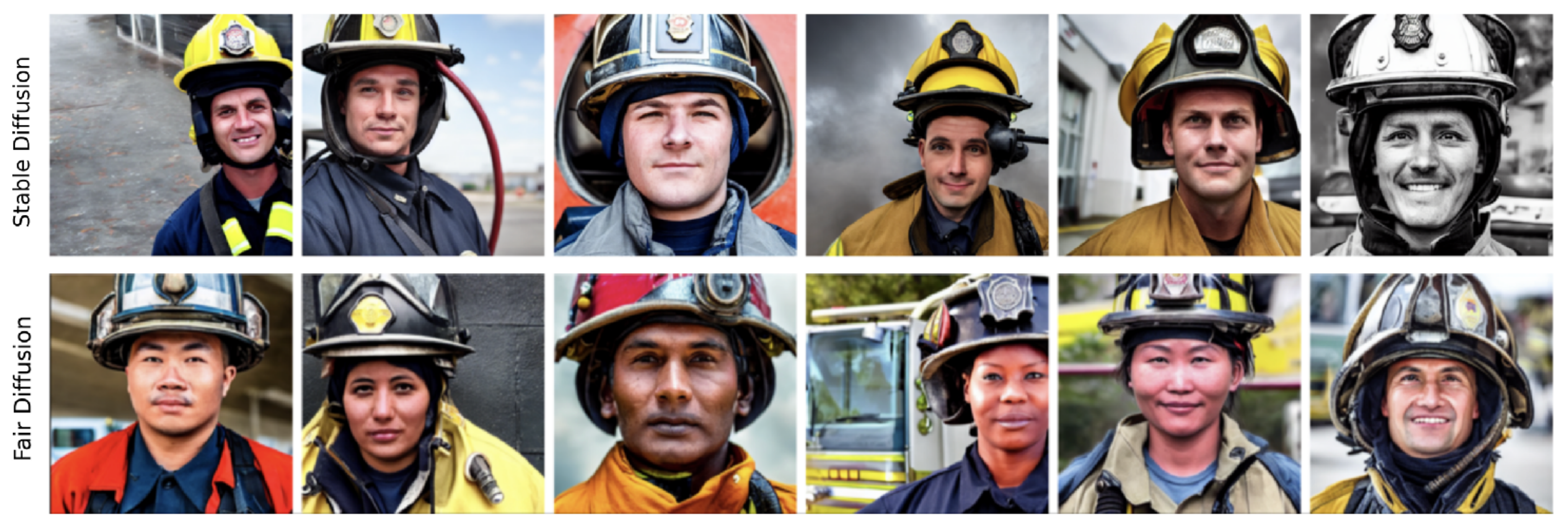}
  \caption{
  Stable Diffusion (top row) runs the risk of lacking diversity in its output (here, e.g., only White male-appearing persons as ``firefighters''). In contrast, \textsc{Fair Diffusion} (bottom row) allows one to introduce fairness---increasing outcome impartiality---according to a user's preferences
  (here, e.g., group identities of ``firefighters'').
  }
  \label{fig:teaser}
\end{figure}

Fig.~\ref{fig:deployment} illustrates how \textsc{Fair Diffusion} can be applied after deployment. As shown in Fig.~\ref{fig:teaser} and discussed later, the default image generation with SD of ``firefighters'' suffers from gender and ethnic biases. With \textsc{Fair Diffusion}, a user provides fair instructions to the model at inference without touching the input prompt. In turn, fair guidance aligns the output with users' fairness notions, e.g., more diverse ``firefighters''.
Here, a lookup table serves as means to identify text prompts requiring fair guidance. It contains user-designed instructions to promote the fair generation of images. This way, \textsc{Fair Diffusion} could be automatized and integrated into APIs. We show further background and full details in the Methods section.



\subsubsection*{Fairness for Diffusion Models}\label{sec:fairness}
As said before, fairness has always been a challenging concept to define \cite{verma18explained,mehrabi21surveybias}. Definitions of fairness and bias are, like many ethical concepts, always controversial, and a variety of definitions exist \cite{binns17fairphilos,verma18explained,mehrabi21surveybias,Hutchinson19history}.
Roughly, fairness can be summarized as the absence of any tendency in favor of a person due to some attribute. 
However, fairness is inherently subjective and suffers from incompleteness. In general, only in very specific and constrained situations, it is possible to satisfy multiple of these fairness notions. In turn, a universal definition is not available \cite{mehrabi21surveybias,fairness_dignum,culture15fairness, culture22fairness}.
We define fairness for \textsc{Fair Diffusion}, in line with closely related work on fairness \cite{xu18fairgan}, as algorithmic fairness for a dataset and model.
\paragraph{Definition 1}\label{def:fair1}
Given a (synthetic) dataset $\mathcal{D}$, fairness or statistical parity is defined as 
\begin{equation}\label{eq:model_fairness}
    P(x,y = 1|a = 1) = P(x,y = 1|a = 0)
\end{equation}
%
%
Here, $y\in \mathcal{Y}$ is the label of a respective data point $x\in \mathcal{X}$, $a$ is a protected attribute and $P$ is a probability. For example, $x$ can be an image with the label $y$ ``firefighter'' and $a$ the protected attribute ``gender''. This fairness definition can be used to evaluate the fairness of a dataset but also of a generative model. Typically, datasets consist of real-world data ($x$) with human labels ($y$). For the evaluation of a generative model, a dataset can be synthetically generated to enable an empirical fairness evaluation. In that case, a data point is obtained through $x=\eta(y)$, where the model $\eta$ is prompted by the user with the desired label $y$. The model $\eta$ can represent any generative downstream task for any input modality (visual, textual, etc.), e.g., $\eta$ can be a generative diffusion model mapping from text (also called prompt $p$) to images. 
In other words, we define a dataset to be fair if Def.~\hyperref[def:fair1]{1} holds, i.e., there is no disproportionate weight in favor of attribute $a$ in the data, and we define a model to be fair if the same holds true for a model's generated output. However, this fairness definition ensures fairness for one binary attribute. In the case of multiple non-binary attributes that may interfere with each other, it becomes more challenging to satisfy them at the same time. Furthermore, this fairness definition requires all attributes to be known, definable, measurable, and separable. We discuss the limitations of this definition in Discussion. And, while there exist multiple definitions of fairness, in this work, we focus on the introduced one to first identify sources of unfairness in DMs and to second evaluate the mitigation of unfairness. 

In line with the presented fairness definition, in Fig.~\ref{fig:deployment}, the concepts $e_1$ and $e_2$ describe the expressions of the (binary) attribute $a$. To ensure statistical parity (Def.~\hyperref[def:fair1]{1}), i.e., that both attribute expressions are equally represented in the model's outcome, we randomly choose the direction of the guidance with equal probability. This results in a uniform probability distribution, assigning the same probability to each expression of an attribute, i.e. $P(a) = \frac{1}{|a|}$.

\section*{Experiments}
\label{sec:experiments}

In this section, we investigate the components of DMs as potential sources of bias on the prominent example of gender occupation biases. Subsequently, we demonstrate their mitigation in the outcome using \textsc{Fair Diffusion}.

For assessing the bias of text-to-image DMs and its mitigation, we inspect the publicly-available diffusion model Stable Diffusion 1.5 (SD \cite{Rombach_2022_CVPR}\footnote{\url{https://huggingface.co/runwayml/stable-diffusion-v1-5}}), its underlying large-scale dataset (LAION-5B \cite{schuhmann2022laion}\footnote{More specifically, the subset LAION-2B(en) 
available at \url{https://laion.ai/blog/laion-5b/}}) and pre-trained model (CLIP \cite{radford2021learning}\footnote{\url{https://github.com/openai/CLIP} with ``ViT-L/14''}). 
For the inspection and generation, we used $p=\text{``A photo of the face of a \{\textit{occ}\}''}$ as a text prompt, where \mbox{$\textit{occ}\in \{\text{``firefighter'', ``teacher'', ``aide'', ...}\}$}.
In terms of fairness, we assume that an equal proportion of female- and male-appearing images is desired as derived from Def.~\hyperref[def:fair1]{1}. 
We present more details on our experimental setup, including measures to quantify DMs' inherited biases and the full experimental protocol in the Methods section.
\begin{figure}[t]
        \centering
        \includegraphics[width=0.96\textwidth]{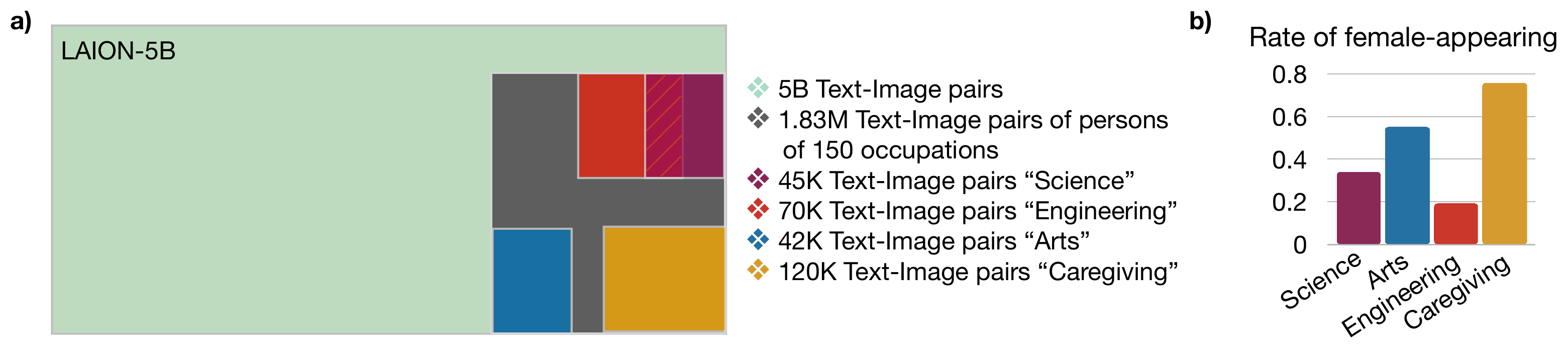}
   \caption{Bias inspection in LAION-5B. \textbf{a)} Proportion of evaluated images. In total, we identified 1.83M images for gender occupation bias. We built four exemplary subsets (``Science'', ``Arts'', ``Engineering'', and ``Caregiving'') of the occupation set to gain first insights into present biases. Some sets overlap (hatched) as the concepts are not disjunct.
   Sizes are only illustrative, and actual numbers are given in the legend. For the final evaluation, we use all 1.8M images. \textbf{b)} Bias evaluation. As shown, the ``Science'' and ``Engineering`` subsets have lower rates of female-appearing persons, while ``Arts'' and ``Caregiving'' have higher rates of female-appearing persons. Consequently, the inspected LAION-5B images represent stereotypical gender occupation biases. (Best viewed in color)}
   \label{fig:dataset-stats}
\end{figure}

\subsection*{Mirrored Unfairness in Stable Diffusion}
We start our empirical study by illustrating the presence of biases in SD and laying the foundation for the subsequent bias mitigation. To this end, we investigated SD's training data, LAION-5B, its pre-trained model, CLIP, and SD's outcome on gender occupation biases.

\begin{table}[t]
        \centering
        \caption{Bias Inspection for CLIP. We examine the iEAT for gender occupation biases. The table shows that such biases are present in CLIP, i.e., male-appearing are considered to be closer to career, science, or engineering, compared to female-appearing who are closer to family, arts, and caregiving. All examples have a high effect size, $d$, and are highly significant, i.e., $p \leq 0.05$ . Furthermore, we evaluated intersectionality biases and found that skin color attributes amplify gender occupation biases.}
        \begin{tabular}{c|cccccc}
         Topic & Target concept & Attribute concept & p ($\pmb{\downarrow}$) & d ($\pmb{\uparrow}$) \\
         \hline
         Gender & Male - Female & Science - Arts & 0.003 & 0.63 \\
         Gender & Male - Female & Engineering - Caregiving & 0.005 & 0.57 \\
         Gender & Male - Female & Career - Family & 0.01 & 0.58 \\
         Ethnicity & White Male - Black Female & Science - Arts & 0.05 & 1.48 \\
         Ethnicity & White Male - Black Female & Engineering - Caregiving & 0.05 & 1.57 \\
         Ethnicity & White Male - Black Female & Career - Family & 0.1 & 0.99 \\
        \end{tabular}
    \label{tab:iEAT_Clip}
\end{table}

\paragraph{Uncovering biases in the data and model foundations of Stable Diffusion.}
To begin with, we evaluated LAION-5B on four subsets of our occupation list (\cf Fig.~\hyperref[fig:dataset-stats]{3a}), where each subset contains images of occupations belonging to one field (science, arts, engineering, and caregiving).\footnote{Selection is explained in App.~D~Tab.~\ref{tab:occupation_group_list}.} Therefore, we classified the images of the subsets for gender to obtain insights into the rate of female-appearing persons, the share of images $\kappa$ (Eq.~\ref{eq:class_inp}) classified as ``female'', as a measure of gender occupation bias. 
Fig.~\hyperref[fig:dataset-stats]{3b} shows that LAION-5B contains several occupation biases.
One can observe that the rate of female-appearing persons is higher for occupation fields like arts or caregiving. On the other hand, the rate is lower for science or engineering. Both demonstrate stereotypical proportions in the dataset.
For inspecting CLIP, we performed a bias association test, the iEAT. In this experiment, we tested the similarity between encoded images of different concepts. In the spirit of Steed \etal \cite{steed21image}, we applied their setup to the CLIP encoder and uncovered similar gender occupation biases (\cf Tab.~\ref{tab:iEAT_Clip}). 
For instance, encoded images of male-appearing persons are closer to engineering-related images than encoded images of female-appearing persons, which are, in turn, closer to caregiving-related images. More interestingly, we found a bias amplification when the association test on gender occupation bias is modified by ethnic attributes. In this case, images of male-appearing people are represented by European-American appearance and images of females by African-American appearance. Accordingly, an intersectionality bias \cite{wang22intersectionality} is present in CLIP too, amplifying the gender occupation bias.
We show more details on this experiment in App.~Tab.~\ref{tab:iEAT_Clip_full}.

\begin{figure}[t]
    \centering
    \includegraphics[width=0.9\linewidth]{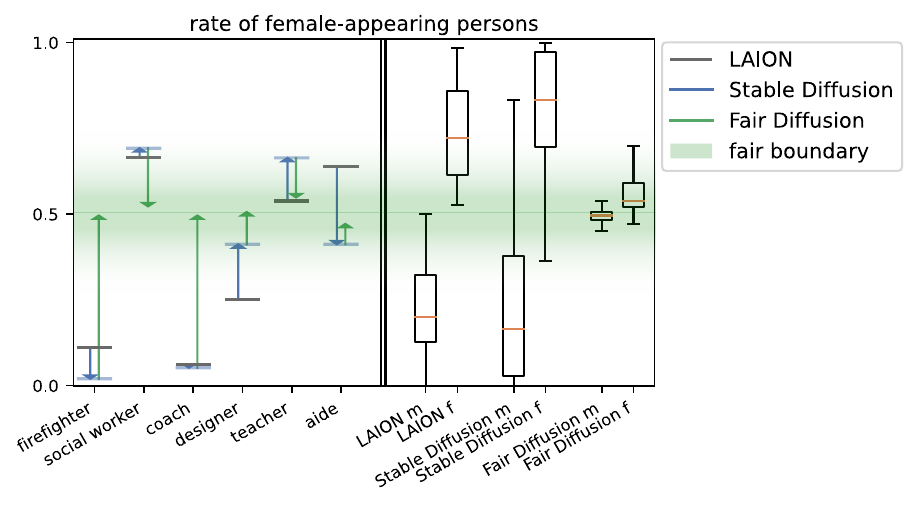}
    \caption{Rate of female-appearing persons for six specific and the per-group median for all 150 occupations. $1.0$ indicates only and $0$ no female persons, whereas $0.5$ indicates $50\%$ are female and $50\%$ male persons.
    Groups f/m represent the subset of more (fe)male-biased occupations. The proportion in the LAION-5B dataset is used as a reference (black vertical bar) for the SD-generated images (blue arrow) which are in turn the reference for \textsc{Fair Diffusion} generated images (ours, green arrow). Here, a proportion toward the middle is preferred over away from the middle, which means a value within the fairness boundary ($50\%\!\pm\!4$). On average, SD-generated images are more biased than the evaluated LAION-5B images, while \textsc{Fair Diffusion} mitigates them toward the fair boundary. For specific occupations, there is no clear tendency for bias amplification between the components. Nonetheless, bias mitigation of SD-generated images, e.g., for ``designer'', does not necessarily yield a fair outcome. In turn, \textsc{Fair Diffusion} shifts the gender proportion within the fair boundary. (Best viewed in color)}
    \label{fig:gender_proportion}
\end{figure}

\paragraph{Uncovering biases in the outcome of Stable Diffusion.}
Next, we examine the third model component, the downstream task and its outcome, for gender occupation biases. Here, we evaluated the generated images from SD. Since SD builds on LAION-5B, we also compared the respective rates in the following to investigate SD's outcome for mirrored unfairness.
This time, we extend the previous investigation to the complete occupation subset of LAION-5B (\cf Fig.~\hyperref[fig:dataset-stats]{3a}).
Fig.~\ref{fig:gender_proportion} depicts the rate of female-appearing persons for six exemplary occupations and its per-group median. One can observe that the SD-generated images (blue lines) also contain clear gender biases for various occupations. For instance, a firefighter or a social worker are significantly affected by bias. The evaluated LAION-5B images (darkgray lines) contain similar biases, providing further evidence for our previous findings. 
However, one can observe a discrepancy in gender occupation biases, e.g., for ``firefighter'', between LAION-5B and SD-generated images. The rate of female-appearing persons is higher for the generated images than for its training data, representing a stronger gender bias. On the other hand, if we look at ``coach'', the gender bias in the SD outcome is on par with LAION-5B. Furthermore, for a designer, the gender bias in the generated images is smaller than in its training data. At the same time, one can also see that there are occupations like ``teacher'', which have nearly no gender bias in LAION-5B (true for overall 5\% of the evaluated occupations); nonetheless, there is a bias in the generated images. Lastly, one can also observe that the gender bias in the generated images for ``aide'' is smaller than in LAION-5B but beyond the fair boundary in the opposite direction. 

\paragraph{Are biases mirrored between LAION-5B and the outcome of Stable Diffusion?}
Overall, we found that LAION-5B's gender biases in the generated images are amplified for 56\%, reflected for 22\%, and mitigated for 22\% of the evaluated occupations\footnote{We denote a reflection if the SD outcome bias corresponds to the LAION-5B bias $\pm4\%$, and an amplification (mitigation) if the bias is farther away from (closer to) the fair boundary.}.
As one can find various behaviors in the SD outcome (amplification, reflection, and mitigation) regarding fairness, we computed a per-group statistic (binary fe/male) to further insight into the overall gender bias in each component. One can observe that the median (orange line) of SD-generated images and the inspected LAION-5B images are distinctly outside the fair boundary. This means LAION-5B and the SD-generated images are unfair according to Def.~\hyperref[def:fair1]{1}. In total, 64 of 150 occupations are more female-biased, while 86 are more male-biased.
More importantly, the median of SD-generated images is farther away from the fair boundary (middle) for both lists (f and m) compared to LAION-5B. 
This provides evidence that the SD output from this experiment is, on average, more biased than the images from LAION-5B. However, one can observe high variance, urging more research in this direction.
Furthermore, attributing the discrepancy in bias to a specific component of the model or aspect of the training procedure is difficult. The shift in bias results from a complex interplay between training data and objective, and CLIP's inherently biased representations, which are in turn influenced by a different training set. \\

\noindent
During our inspection, we found biases and unfairness in each component of the SD pipeline: in the LAION-5B dataset, in the CLIP encoder, and in the generated images. At the same time, the biases are not simply mirrored between LAION-5B and SD's outcome and do not show a clear tendency.

\subsection*{Instructing on Fairness with \textsc{Fair Diffusion}}
After discovering several biases in SD's components, we turn to mitigate them. Since the interplay between SD's components is complex, debiasing them is a challenging task. Unlike related work on debiasing, we instruct a model on fairness at the deployment stage. 
Next, we evaluate the guidance toward fairness of text-to-image generations. 

\paragraph{Setting up \textsc{Fair Diffusion}.}
In contrast to the image generation with default SD, we further conditioned the image generation with \textsc{Fair Diffusion}. To this end, we steered the image generation toward ``female person'' ($e_1$) and away from ``male person'' ($e_2$) and randomly switched the direction toward male-appearing and away from female-appearing with a 50\% chance.\footnote{We would like to emphasize again the limitation of binary gender in this setup (\cf Discussion).} This way, we utilize the concepts encoded in a DM to simultaneously suppress one and reinforce the other, with alternating directions. Due to this approach, 50\% of the images should contain (fe)male-appearing generated persons.\footnote{We applied the chosen guidance to the image generation regardless of its outcome without guidance or any biases present in LAION-5B.}
For an apt comparison, we applied \textsc{Fair Diffusion} to SD-generated images, i.e., we re-generated an image with the same seed and parameters and included the additional conditioning (fair guidance $\gamma$) for gender.

\paragraph{Mitigating gender bias in the outcome of Stable Diffusion.}
Fig.~\ref{fig:gender_proportion} demonstrates \textsc{Fair Diffusion}'s performance (green arrow) in mitigating gender occupation biases detected in Stable Diffusion (blue bar). 
Looking at our six exemplary cases, one can observe a shift of the gender proportion to the inside of the fair boundary, no matter in which direction the bias was previously in. 
Furthermore, the biases can be addressed with the \textsc{Fair Diffusion} strategy, whether they are present in LAION-5B or the SD outcome and whether the SD outcome amplifies, reflects, or mitigates LAION-5B biases. For example, though the proportion in SD-generated images for ``designer'' is less biased than in LAION-5B, it is still not within the fair boundary. In turn, \textsc{Fair Diffusion} mitigates the bias further and shifts the gender proportion within the fair boundary.
Moreover, the per-group (m/f) median (\cf Methods) is within the fair boundary, so \textsc{Fair Diffusion} successfully reduced unfair occupation proportions. Hence, on average, \textsc{Fair Diffusion} achieves fairness according to Def.~\hyperref[def:fair1]{1}, i.e., for the model outcome. 
However, we can see that there remains variance in the generated images for some occupations. This can generally be due to the non-binary nature of gender, and gender is also not to be determined simply based on outward appearance. Moreover, we could identify some outliers, e.g., images of ``dishwasher'' were generally difficult to generate but also difficult to edit for gender, as it does describe not only an occupation but also a cleaning device. When searching LAION-5B\footnote{\url{https://knn5.laion.ai/}} for ``a photo of the face of a dishwasher'', we also mainly found images of the cleaning device and no humans. So, we assume this to be an artifact due to the ambiguity of ``face of a dishwasher''. For a broader evaluation, in particular, on the design of the editing prompts, we refer to App.~\hyperref[sec:dif-setups]{C}.

\begin{figure}[t]
    \centering
    \includegraphics[width=0.95\linewidth]{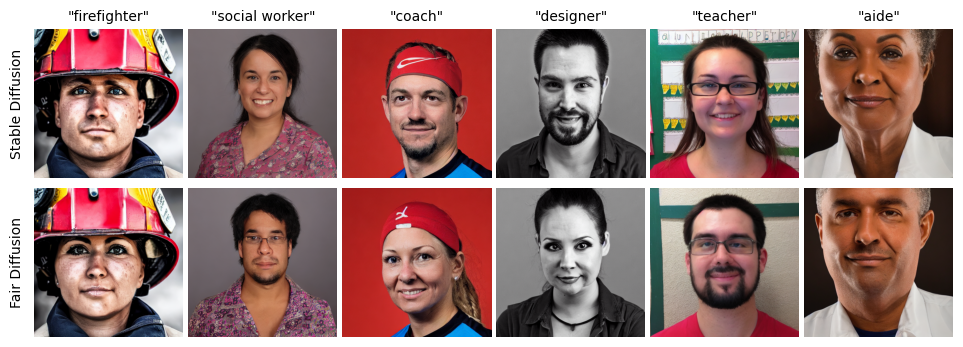}
    \caption{Generated Images with SD (top row) and \textsc{Fair Diffusion} (bottom row) for different occupations. The images are generated with the prompt ``A photo of the face of a \{$occ$\}'', in which each column name represents the used occupation ($occ$). For generated images of female-appearing persons, we applied fair guidance with -``female person'' +``male person'' and vice versa for male-appearing persons.
    One can observe that \textsc{Fair Diffusion} changes the typical gender appearance for each occupation image while keeping the residual (occupation-related) features present.}
    \label{fig:occupation_example}
\end{figure}

Apart from our quantitative analysis, we also show qualitative results on our bias mitigation approach in the following. In Fig.~\ref{fig:occupation_example}, one can observe a shift in outward appearance. The top row shows images generated with SD and the prompt $p$ for each of our six exemplary occupations. As our inspection experiments already demonstrated, there is mostly a strong bias toward one gender for certain occupations. 
In contrast, the images generated with \textsc{Fair Diffusion} (bottom row) shift the gender appearance toward the other gender appearance and by that ultimately toward a more diverse output. Notably, the overall image composition remains the same, with only minor changes to the rest of the image. 

\paragraph{Approaching bias beyond gender.}
Another step toward a fairer model outcome is to go beyond editing gender appearance. In Fig.~\ref{fig:teaser}, we used \textsc{Fair Diffusion} to edit multiple features of firefighters' outward appearance. The result is a more diverse output of facial features in terms of gender, skin tone, and ethnicity. This is due to the capability of \textsc{Fair Diffusion} to modify image features according to user preferences in isolation. This way, we can remove biases and increase outcome impartiality beyond gender. In App.~\hyperref[sec:add-fadi-imgs]{A}, we show further possible approaches to biases beyond gender, e.g., heteronormativity and age discrimination. \\

In summary, it is difficult to create a model with fair outcomes in all aspects. Here, we evaluated instructing text-to-image models to approach outcome impartiality. Our empirical results on \textsc{Fair Diffusion} demonstrated its potential as a reliable approach. However, we emphasize the interplay between processing training data and fairness in representations as well as the outcome. 
We envision a future with models that are able to generate a more diverse outcome---hand in hand with the user who steers and controls the generation process.

\section*{Discussion}\label{sec:discussion}
We have found several severe biases in all components of the SD pipeline and introduced \textsc{Fair Diffusion} to mitigate these biases. Still, we need to delve deeper into some of the insights gained and use this section to discuss them in greater detail. In particular, we touch upon opportunities for society.

\paragraph{Shifting the debiasing paradigm.} This work provides food for thought on current debiasing techniques, mostly focused on dataset curation and in-process bias prevention, shifting the focus to the deployment stage. As Nichol \etal \cite{nichol2022glide} showed, curating datasets by filtering has drawbacks, such as persisting biases and worse generalization capabilities. In turn, \textsc{Fair Diffusion} operates at the deployment stage enabling fair outcomes according to Def.~\hyperref[def:fair1]{1}. On the other hand, our inspection also showed that the evaluated images from LAION-5B are, on average, still remarkably affected by bias. Consequently, debiasing it might remain important. Ultimately, we believe debiasing all components might be necessary to increase fairness in generative models further. Nevertheless, as long as this is not available, especially for end-users, we demonstrated that the text interface of DMs enables instructions as an easy-to-use technique that can be immediately deployed to mitigate biases in the outcome of current image generation models.
This does not introduce an entirely \textit{fair} model but instead a way to control unfair models to increase fairness. In the presented version, \textsc{Fair Diffusion} is applied regardless of, e.g., existing biases in LAION-5B.
While \textsc{Fair Diffusion} enables a user to take control of fairness, a future vision for even fairer models is automated detection of unfairness. If biased concepts are known beforehand, a user could supply \textsc{Fair Diffusion} with these in order to take action without a user actively intervening in the moment of generation. For example, the lookup table and instructions in Fig.~\ref{fig:deployment} could be filled out beforehand.

\paragraph{Is the model finally fair?} As shown, \textsc{Fair Diffusion} promotes fairness according to Def.~\hyperref[def:fair1]{1}. However, as discussed before, fairness is inherently incomplete, such that the setup used in this work does not account for other fairness definitions.

For example, it realizes fairness according to Def.~\hyperref[def:fair1]{1} in the model outcome but cannot in the dataset. To achieve such fairness, the dataset has to be filtered or modified so that no biases remain. 
Still, \textsc{Fair Diffusion} can be used to realize different notions of fairness. If the desired output proportion differs from an equal proportion (50/50 for binary attributes), a user can realize this easily by setting a different edit probability $P$ that is non-uniform (e.g., to 70/30 for binary attributes). This may be used for a fairness definition that simply reflects the occupation proportions of current society, i.e., utilizing a country's occupation statistic a user lives in. This further illustrates that fairness is versatile, and \textsc{Fair Diffusion} is easily adaptable to these different notions.

\paragraph{Binary gender labeling.} 
We acknowledge the limited representation of gender in this study.
Current automated measures
treat gender as a binary-valued attribute, which it is not \cite{wickham2023gender}. 
Due to the lack of tools that classify beyond binary gender, an empirical evaluation at a large scale remains limited to binary-valued gender evaluation. 
Similarly, we assume encoded concepts in Stable Diffusion to be limited to binary gender terms. Consequently, we observed non-binary fairness instructions to result in fragile behavior (\cf App.~\hyperref[sec:dif-setups]{C}).
Although current diffusion models have inherent limitations, \textsc{Fair Diffusion} builds on them to make a first step toward fairness by mitigating, e.g., gender-biased image generation. We advocate for research on encoding gender as non-binary in generative and predictive models.

\paragraph{Technical limitations.}
Furthermore, we also want to touch upon some technical limitations of this work. First, when evaluating the LAION-5B dataset, we observed that several images are stock photos. This is a reminder that LAION-5B is a web-crawled dataset that does not represent reality, nor does it reflect the internet in its entirety. 
Second, we use the CLIP encoder for searching LAION-5B. However, as demonstrated, CLIP is inherently biased \cite{wolfe22americanwhite,wolfe22markedness}, which may affect the search results. Therefore, the resulting images are not entirely disentangled from this confounding factor. However, there are barely any alternatives, as manual labeling is biased too and infeasible due to a large amount of data, and other automated approaches will suffer from imprecision too. We empirically chose the threshold $\rho$ to be low enough to counteract this behavior, so images of disadvantaged genders will be included in the search result.
Third, the gender classification results rely on a pre-trained classifier, FairFace \cite{fairface}. As said, the classifier is an inherent limitation for classifying (binary) gender. Furthermore, we cannot guarantee that this classifier is bias-free and hence promote investigating its function and alternative ways. Yet, it seems to be the best available choice for an automated evaluation. Moreover, FairFace and the other limitations are only relevant to evaluate \textsc{Fair Diffusion}, while the strategy itself is independent of, e.g., the used classifier for evaluating.
Lastly, \textsc{Fair Diffusion} builds on \textsc{Sega} and inherits its constraints. The concepts (biases) to be removed must be describable, i.e., either with natural language or with images. However, this is inherent to all currently available approaches that enable editing generative DMs. \textsc{Fair Diffusion} is agnostic to that, and \textsc{Sega} can, in principle, be replaced with other editing techniques.

\paragraph{Beyond text-guided fairness.} 
\textsc{Fair Diffusion} currently operates on the textual interface to steer image generation. However, it is not limited to the textual modality.  \textsc{Fair Diffusion} adds fair guidance to the image generation according to edit encodings $c_{e_i}$. Currently, CLIP's text encoder embedded the edit prompt $e_i$ to $c_{e_i}$. The way the encoding is obtained can go beyond (English) natural language. \textsc{Fair Diffusion} could be extended with other approaches like AltCLIP \cite{chen22altclip} for multilingual encodings, 
Textual Inversion \cite{Gal22Textinversion} for visual encodings, or MultiFusion \cite{bellagente2023multifusion} for multimodal (text and image) encodings. This variety of approaches offers versatile interfaces for fair guidance with \textsc{Fair Diffusion}.

\paragraph{The challenges of user interaction.} While human interaction has generally proven to be helpful \cite{teso2019xil,Ouyang2022TrainingLM,friedrich2022XIL_typo, friedrich22rit}, at the same time, certain dangers can arise. For example, a user in control with malicious intentions could target the model to misuse it. Like many other pieces of research, \textsc{Fair Diffusion} faces the dual-use dilemma. The strategy can be used in an adversarial manner as well, such that biased outcomes of generative models can be further amplified, and its diversity decreased. Hence, further detection mechanisms for malicious interaction are required. This is an active research topic \cite{avoidtrolls_ju} that needs consideration when using human interaction. \\

Our work is of greater relevance as it offers the opportunity to immediately promote fairness in many real-world applications. As image generation models become increasingly popular and integrated into our lives, fairness must be kept in mind. DMs come into play even in high-stakes applications such as medicine and drug development \cite{Watson2022protein}. These models are also used in other areas, such as advertisement or design\footnote{\url{https://www.cosmopolitan.com/lifestyle/a40314356/dall-e-2-artificial-intelligence-cover/}}. Imagine a firefighter advertisement\footnote{For example, stock images are already being generated using diffusion models (Shutterstock, \url{https://www.shutterstock.com/press/20435})} containing people from Fig.~\ref{fig:teaser} top or bottom row only. This way, generative models can have a crucial impact on societies and how we include and value diversity in them. 
Furthermore, \textsc{Fair Diffusion} can make another step toward fairness in society. This work focuses on a specific definition of fairness for evaluation purposes. However, the way such a tool is used also has a political dimension beyond research. Sometimes, the goal is not to achieve an equal outcome for each attribute. Temporarily over-representing a certain attribute, e.g., in advertisements, can be desired as it can promote awareness and transparency for bias and discrimination concerning this attribute. Or, current over-representations can be gradually reduced, to slowly habituate new proportions in a society. The pathway toward an ideal discrimination-free world may take measures that might contradict the fairness definition used in this work (Def.~\hyperref[def:fair1]{1}) but align with other fairness definitions \cite{mehrabi21surveybias}. Hence, our approach facilitates flexible outcome proportions, which, in turn, enables over-representation or any other proportion.
However, it remains an open question, what the new world (à la Aldous Huxley) should look like that an AI system reflects. We do not argue for any specific proportion or promote any specific political direction. Instead, we provide a strategy that can be used by society and politics immediately with ease for purposes that can ultimately promote a fairer depiction of society. Therefore, the overall goal might not be a fair tool itself, as it is rather a means to an end, but to use it in a way that promotes a fairer society without discrimination.

\section*{Conclusion}
We introduced \textsc{Fair Diffusion} and demonstrated that it can instruct generative text-to-image models in terms of fairness. To measure fairness, we explored the publicly available large-scale training dataset of Stable Diffusion (LAION-5B) and further applied iEAT on its underlying pre-trained representation encoder (CLIP). Both show severe gender and racial biases mirrored in the downstream diffusion model. However, its textual interface and advanced steering approaches provide the necessary control to instruct the generative model on fairness, as our extensive evaluation demonstrates. Specifically, we show how to shift the bias in generated images in any direction yielding arbitrary proportions for, e.g., gender and race. In this way, our method prevents diffusion models from implicitly and unintentionally reflecting or even amplifying biases.
Based on our findings, we strongly advise careful usage of such models. However, we also envision easily accessible generative models as a tool to amplify fairness, i.e., itself introducing syntactic biases---compared to real-world distributions---into realistic images. This enables media to display various genders in (fe)male over-represented occupations motivating younger people to follow their interests despite societal biases \cite{shaw2010identity, caswell2017to, rawan2019why}.

An exciting avenue for future work is disentangling the components to pinpoint the sources of bias in the model. Furthermore, it is interesting to compare different generative diffusion models for fairness. In addition, this work can be extended to image-to-image diffusion, facilitating the editing of real-world images rather than just generated ones.
Lastly, \textsc{Fair Diffusion} can be easily integrated into any real-world diffusion application, mitigating unfair image generation or even amplifying fairness.

\section*{Methods}
In this section, we introduce the underlying components of \textsc{Fair Diffusion}, measures to evaluate fairness in diffusion models and a detailed experimental protocol.

\begin{figure}[t]
    \centering
    \includegraphics[width=0.75\linewidth]{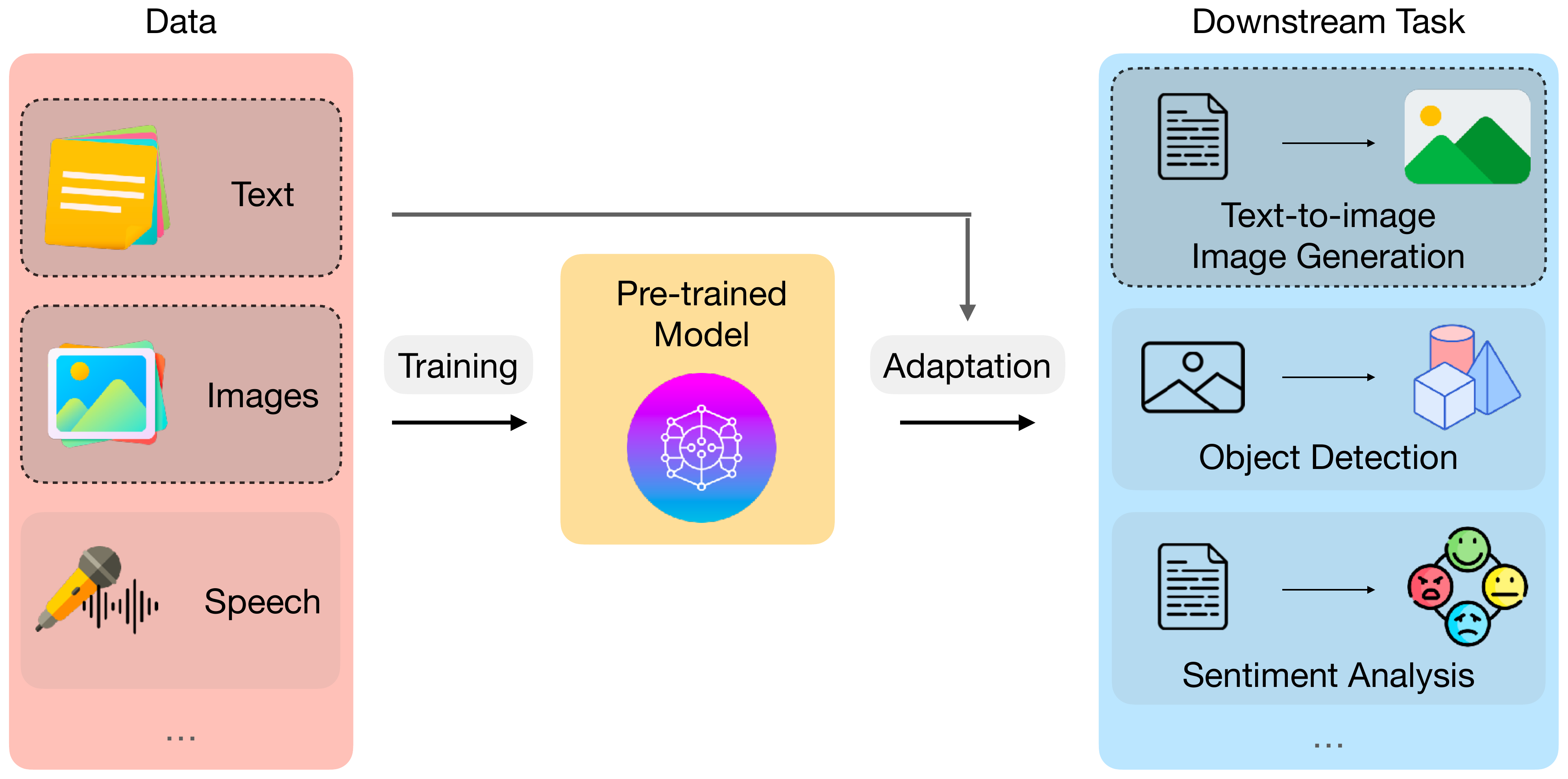}
    \caption{Setup of large-scale AI models. Many recent AI systems are centered around a pre-trained model \cite{Bommasani21foundationmodels}. For one, this model is trained on large-scale data, often multimodal. On the other hand, it is adapted to a downstream task, e.g., by fine-tuning. This work focuses on DMs, and for many of them, CLIP serves as the pre-trained model, which is trained on text and image data (dashed boxes, left). It is adapted for the downstream task such that only its text encoder is integrated into the DM to generate images from this text (dashed box, right). In turn, the DM is adapted to the downstream task by fine-tuning on task-specific data. (Best viewed in color)}
    \label{fig:pipe}
\end{figure}

\subsubsection*{Text-guided Image Generation}\label{sec:diffusion}
As visualized in Fig.~\ref{fig:pipe}, 
many recent models, like DMs, rely on large-scale training datasets as well as large-scale pre-trained models. These requirements are necessary to perform well in text-to-image generation tasks and generalize over multiple domains. The information transfer from the pre-trained model and the downstream training data helps these models achieve remarkable performance. However, combining both also increases the risk of introducing biases into the models, as we demonstrated in our experiments. 

The underlying intuition of DMs for image generation is as follows: 
the generation starts from random noise $\mathbf{z}$, and the model predicts an estimate of this noise $\Tilde{\epsilon}$. Subtracting this estimate from the initial value results in a high-fidelity image $x = \mathbf{z} - \Tilde{\epsilon}$ without any noise. 
Since this is an extremely hard problem, multiple steps $T$ are applied iteratively, each subtracting a small amount ($\Tilde{\epsilon_t}$) of noise, approximating $\Tilde{\epsilon}$. For text-to-image generation, the model's $\Tilde{\epsilon}$-prediction starts of from random noise $\mathbf{z}_t$ and is conditioned on encoding $\mathbf{c}_p$ of text-prompt $p$ and results in an image faithful to that prompt.
This results in:
\begin{align}\label{eqn:dm_eqn}
    \Tilde{\epsilon_t} &= \text{DM}_{\theta}(\mathbf{z}_t, \mathbf{c}_p) = \Tilde{\epsilon_\theta}(\mathbf{z}_t, \mathbf{c}_p) \\
    \mathbf{z}_{t+1} &= \mathbf{z}_t - \Tilde{\epsilon_t} 
\end{align}
Consequently, the final image $x$ is equivalent to the last denoising step $x = \mathbf{z}_T$.
In the following, we use $\Tilde{\epsilon_\theta}$ instead of DM$_\theta$ for better readability, in line with previous work.

The textual interface, i.e., text conditioning, is realized through classifier-free guidance \cite{ho2022classifier}, the standard technique for current diffusion models. 
More specifically, during image generation an unconditioned $\Tilde{\epsilon_\theta}$-prediction, $\Tilde{\epsilon_\theta}(\mathbf{z}_t)$, is pushed in the direction of the conditioned $\Tilde{\epsilon_\theta}(\mathbf{z}_t, \mathbf{c}_p)$ to yield an image faithful to prompt $p$. For the interested reader, more details on the general function of diffusion models can be found at \cite{luo2022understanding} or on other websites\footnote{E.g. \url{https://en.wikipedia.org/wiki/Diffusion_model}}.
As we will show in the following, Brack \etal \cite{brack2023Sega} further improved on this guidance to take control and steer the generated image by allowing for multiple additional instructions during the image generation. 

\subsubsection*{Bias Mitigation}
\label{sec:related}
Recently, many approaches have been proposed to create models with fairness in mind. For large-scale models, these methods can be categorized with respect to three paradigms: (i) pre-processing the training data to remove bias before learning \cite{yang20balancing,Prabhu2020LargeID,schramowski2022can,xu18fairgan,Sattigeri19fairngan,nichol2022glide}, (ii) enforcing fairness during training by introducing constraints on the learning objective \cite{edwards15adversary,zhang18adversarialfairness,Li_2022_ECCV,berg-etal-2022-prompt}, and (iii) post-processing approaches to modify the model outcome at the deployment stage \cite{schramowski2022safe,bansal2022diversify}. 

For DMs, Nichol~\etal \cite{nichol2022glide} filtered the data prior to training in order to mitigate bias through the removal of data related to certain concepts. However, they observed the filtered model to continue exhibiting bias while encountering adverse effects such as a loss in generalization ability. 
These results highlight again that creating a completely bias-free dataset is generally infeasible.
Additionally, different definitions of fairness would each require a dedicated dataset and respective model tailored to the targeted fairness characteristics.
This contradicts a major principle of large-scale pre-training, i.e., training one model on only one (large) dataset and subsequently using it for downstream tasks.
Hence, data pre-processing alone does not provide an apt solution for mitigating biases.
 
In contrast, this work targets the (post-process) deployment stage of DMs. 
Fortunately, large-scale models are not at the mercy of under-curated data. Schramowski \etal \cite{schramowski2022safe} demonstrated that representations learned during pre-training can be exploited and leveraged to suppress unwanted and inappropriate behavior in the downstream task. Whereas their work focused on suppressing inappropriate (e.g., pornographic) content, we used a similar approach for fair outcomes, giving power to the individual user. This allows them to instruct the model on their individual definition of fairness. Other works have already shown that user instructions are an essential component for machine learning models to enable user alignment \cite{schramowski2022large,Ouyang2022TrainingLM,friedrich22rit}, trust \cite{teso2019xil}, and overall model performance \cite{friedrich2022XIL_typo}.

\paragraph{Fair guidance.}
As discussed before, \textsc{Fair Diffusion} can be implemented with any image editing approach. In this work, we chose \textsc{Sega} \cite{brack2023Sega} as it offers a powerful interface to conduct image editing. \textsc{Sega} extends on the principles of classifier-free guidance, presented in previous section about text-guided image generation. In addition to the text prompt $p$ (and its guidance scale $s_g$), a user can provide additional textual concept descriptions $e_i$ (with their own scale and direction). This extends the previous noise estimation for image generation from $\mathbf{\Bar{\epsilon}}_\theta(\mathbf{z}_t, \mathbf{c}_p)$ (Eq.~\ref{eqn:dm_eqn}) to $\mathbf{\Bar{\epsilon}}_\theta(\mathbf{z}_t, \mathbf{c}_p, \mathbf{c}_e)$ (Eq.~\ref{eqn:fairguidance}). In this way, the initially unconditioned image generation is additionally conditioned on the input prompt (classifier-free guidance) and the user instructions, e.g. on fairness (fair guidance $\gamma$).
In particular, multiple concepts can be combined arbitrarily and either increased or decreased during the image generation. This way, we can also realize more complex changes by e.g. guiding toward the encoding $c_{e_1}$ of one concept $e_1$ and simultaneously away from $c_{e_2}$ with text $e_2$, e.g., its alleged opposite. 
Overall, the resulting $\epsilon$-estimate, $\Bar{\epsilon_\theta}$, can be written as:

\begin{equation}\label{eqn:fairguidance}
\mathbf{\Bar{\epsilon}}_\theta(\mathbf{z}_t, \mathbf{c}_p, \mathbf{c}_e)=        \underbrace{ \mathbf{\epsilon}_\theta(\mathbf{z}_t) + s_g \big(\mathbf{\epsilon}_\theta(\mathbf{z}_t, \mathbf{c}_p) - \mathbf{\epsilon}_\theta(\mathbf{z}_t)\big)}_{\text{classifier-free guidance}} + \underbrace{\gamma(\mathbf{z}_t, \mathbf{c}_e)}_{\text{fair guidance}}
\end{equation}
More technical details on this technique can be found by Ho \etal \cite{ho2022classifier} and Brack \etal \cite{brack2023Sega}.

In contrast to this implementation of fair guidance, other approaches such as simple prompt engineering are limited in their abilities \cite{bansal2022diversify} and often change the entire image composition. On the other hand, \textsc{Sega} enables editing of dedicated features in isolation. Furthermore, changing gender appearance by inserting gender-related words into the input prompt is a non-trivial task for machines and requires language understanding. Determining the right position for the edit word(s) is challenging, especially if biases in the prompt are present only implicitly. \textsc{Sega} implements the challenging integration, e.g. of negations and conditions, in a sentence in a technical way. This way, it needs no such language understanding while enabling the same capacity as the edit instruction, i.e. the fair guidance) is independent of the input prompt. We illustrate this issue in App.~Fig.~\ref{fig:baseline_comparison}.

\subsection*{Measuring Fairness in the Components of Diffusion Models}\label{sec:meth_inspect}
In previous sections, we showed that DMs are built around various components (\cf Fig.~\ref{fig:pipe}) and each can be affected by bias. Namely, bias in the datasets, the learned representations, and its reflection in the outcome. Next, we describe the measures used to quantify and track biases across all three components.

\paragraph{Datasets.}
The first potential source of bias, according to the general model setup (\cf Fig.~\ref{fig:pipe}), is the dataset. 
Given a potentially biased attribute, e.g., gender, we investigate its co-occurrence with a target attribute such as occupation. 
If, for example, the proportion of genders within all samples of an occupation is not in line with the fairness definition (e.g. Def.~\hyperref[def:fair1]{1}), we have identified a source of bias already emanating from the dataset. 
This proportion also serves as a reference to investigate whether the model outcome reflects, amplifies, or mitigates such a bias.
If the investigated dataset has no pre-existing labels for the attribute(s) of interest, they have to be derived first. 
However, this is generally a non-trivial task and cannot easily be transferred between different domains.
For vision-language tasks, a sensible approach is to employ a multimodal model capable of computing text-image similarity. We identified relevant images $\mathcal{R}$ in the dataset by computing their similarity to a textual description $p$ of the target concept \cite{radford2021learning}. Along these lines, we obtain the label at the same time, as textual description $p$ corresponds to label $y$.

In this work, we selected images aligned with description $p$ by filtering the entire dataset with an empirically-determined threshold~$\rho$:
\begin{equation}
    \mathcal{R} = \{i \,\, | \,\, sim(i, p) > \rho \,\, \text{and} \,\, i \in \mathcal{I} \}
\end{equation}
where $\mathcal{I}$ denotes the set of images from the dataset.
Next, we used a pre-trained classifier, $\kappa$, to determine the (missing) label for the protected attribute under investigation. Consequently, we obtained each label $a_r$ for image $r\in\mathcal{R}$ with:
\begin{equation}
    a_r = \kappa(r) 
    \label{eq:class_inp}
\end{equation}

\paragraph{Representations.}
Second, we investigated the bias of learned representations using the image Embedding Association Test (iEAT) \cite{steed21image}.
Intuitively, iEAT tests for statistically significant associations between sets of representations, e.g., encoded images. These consist of two attribute sets $A$ and $B$ and two target sets $K$ and $L$. A common example is target images of female-appearing people $L$ and male-appearing people $K$ compared against images related to career $A$ and family $B$. This way, a biased model may associate the images of male-appearing people closer to ``career'' than to ``family'' and vice versa.
Formally, the test statistic can be computed as:
\begin{equation}
    s(K,L,A,B) = \sum_{k\in K}s(k,A,B) - \sum_{l\in L}s(l,A,B)
\end{equation}
where
\begin{equation}
    s(w,A,B) = \text{mean}_{a\in A}cos(w,a) - \text{mean}_{b\in B}cos(w,b)
\end{equation}
 This way, $s(w,A,B)$ computes the association of an encoded image $w$ with the attributes ($a$ and $b$) and eventually the differential association of the encoded target images with the attributes.
We assess the statistical significance by computing the one-sided $p$-value along with the effect size $d$ as:
\begin{align}
    p&=Pr_i[s(K_i,L_i,A,B)>s(K,L,A,B)] \\
    d&=\frac{mean_{k\in K}s(k,A,B) - mean_{l\in L}s(l,A,B)}{\sigma_{w\in K \cup L}(s(w,A,B))}
\end{align}
where $\sigma$ denotes the standard deviation.

\paragraph{Outcome.}
The third source of bias we inspected is the downstream task approximated by its outcome. 
The modality of the outcome generally depends on the type of model and task. Here, we evaluated images generated by Stable Diffusion. 
The procedure to inspect these images for bias is similar to the dataset inspection: a synthetic image dataset is created, attribute correlations in it are calculated which are in turn evaluated for fairness, e.g. according to Def.~\hyperref[def:fair1]{1}.
To investigate potential bias transfer between the training data and outcome (mitigation, reflection or amplification), we generated images using the same text prompt $p$ used for searching the dataset. 
Similarly, we again used the same classifier $\kappa$ (Eq.~\ref{eq:class_inp}) to determine label $a_g$ for the protected attribute in the generated images ($g\in\mathcal{G}$) with $a_g = \kappa(g)$.

\subsection*{Experimental Protocol}
For the inspection, we created a new subset of LAION-5B with over 1.8 million images displaying humans with recognizable faces and in recognizable occupations (\cf Fig.~\hyperref[fig:dataset-stats]{4a}) and generated over 37,000 images each with SD and 
\textsc{Fair Diffusion}, respectively. In total, we evaluated more than two million images for gender occupation biases.
Our instruction tool is built around \textsc{Sega}\footnote{Code publicly available at \url{https://github.com/ml-research/semantic-image-editing}} to edit images and guide the image generation toward fairer outcomes and employed FairFace\footnote{Code publicly available at \url{https://github.com/joojs/fairface}} \cite{fairface} as $\kappa$ to classify the protected attribute, i.e. facial (gender) attributes. Yet, \textsc{Fair Diffusion} can facilitate other image editing and classifying tools, too.

\paragraph{Prompt design.} We employed CLIP to identify relevant images in LAION-5B---i.e., depicting people in recognizable occupations---and computed text-image similarities between LAION-5B images and a text prompt representing an occupation. To this end, we used $p=\text{``A photo of the face of a \{\textit{occ}\}''}$ as a text prompt and empirically determined a similarity threshold $\rho=0.27$. We also used this prompt to generate images with SD, where \mbox{$\textit{occ}\in \{\text{``firefighter'', ``teacher'', ``aide'', ...}\}$}. The whole list consists of over 150 different occupations\footnote{Taken from \url{https://huggingface.co/spaces/society-ethics/DiffusionBiasExplorer}} and we generated 250 images for each occupation prompt.

\paragraph{Fairness assumptions.} We made the assumption that an equal proportion of female- and male-appearing images is desired as derived from Def.~\hyperref[def:fair1]{1}. However, our evaluation is limited by current classifiers (like FairFace) facilitating only binary-valued gender classification, whereas gender is clearly non-binary (extensively examined in Discussion). Interestingly, \textsc{Fair Diffusion} can, in principle, be applied to non-binary gender identities and may also be used to realize different target distributions.
Moreover, we employed a fair boundary (i.e., allow for a deviation of $\pm4\%$) to soften this theoretical assumption. This way, we try to account for natural non-perfect binarity, i.e. the continuous spectrum of gender with its diversity and non-equal birth rate and world population.

\paragraph{Statistical measures.} We computed a per-group statistic (binary fe/male appearing) to further insight into the overall gender occupation bias in each component. Therefore, we divided the list of occupations into f and m, where the f-group denotes more female-biased occupations and the m-group otherwise. If the rate of female-appearing persons in LAION is $>\!0.5$ we use f and otherwise m, respectively. Subsequently, we evaluate these lists for each component and generate respective box plots. Without this group distinction, the average bias lies within the fair boundary (although the box plot shows high variance) as there are strong biases in both directions, which cancel each other out in an overall mean computation.

\section*{Acknowledgments}
This work benefited from the ICT-48 Network of AI Research Excellence Center ``TAILOR'' (EU Horizon 2020, GA No 952215), the Hessian research priority program LOEWE within the project WhiteBox, and the Hessian Ministry of Higher Education, Research and the Arts (HMWK) cluster projects ``The Adaptive Mind'' and ``The Third Wave of AI'', and from the German Center for Artificial Intelligence (DFKI) project ``SAINT''.

\clearpage

\bibliographystyle{bst/sn-vancouver}
\bibliography{references}

\appendix
\section*{Appendix}
\subsection*{A \quad Further Applications and Results with \textsc{Fair Diffusion}}
\label{sec:add-fadi-imgs}
\begin{figure}[b]
    \centering
    \includegraphics[width=0.9\linewidth]{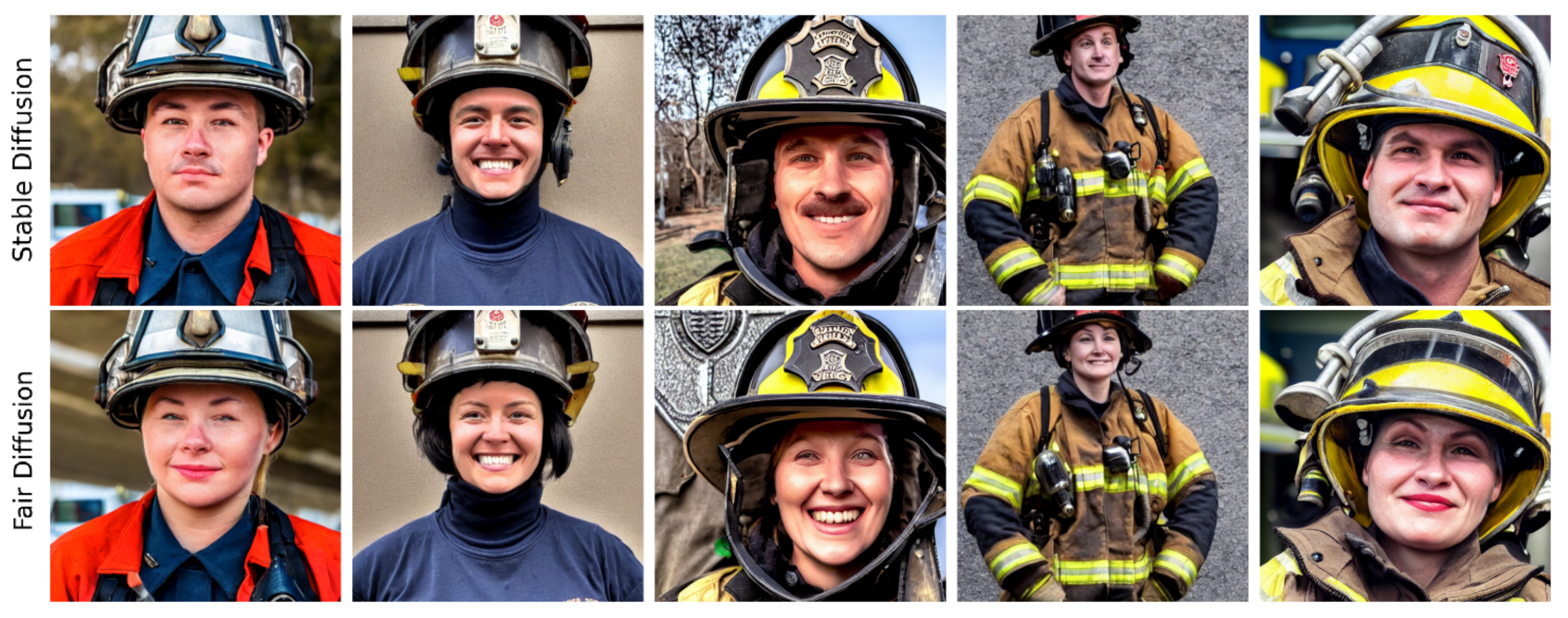}
    \caption{Generated Images with SD (top row) and \textsc{Fair Diffusion} (bottom row) for occupation ``firefighter''. The images are generated with the prompt ``A photo of the face of a firefighter''. We applied fair guidance with -``male person'' + ``female person''. One can observe that \textsc{Fair Diffusion} changes the typical gender appearance for each occupation image while keeping the residual (occupation-related) features present.}
    \label{fig:firefighter_example}
\end{figure}

Apart from the results shown in the main text, we also generated more images, to provide insights into \textsc{Fair Diffusion}. Fig.~\ref{fig:firefighter_example} shows again that generated images of ``firefighters'' by default SD (top row) are strongly male-biased. In contrast, \textsc{Fair Diffusion} changes the outward appearance towards female-appearing ``firefighters''. More interestingly, the changes in gender appearance do not change the overall image composition and the occupation remains identifiable. We regard this as a very powerful property of our approach.

\begin{figure}[b]
    \centering
    \includegraphics[width=0.9\linewidth]{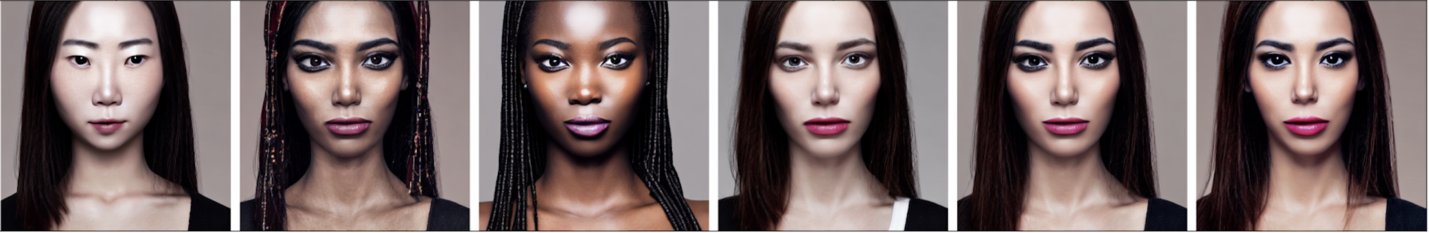}
    \caption{Generated images with \textsc{Fair Diffusion} for ``A photo of a woman''. The images are instructed with +``Asian'', +``Indian'', +``African'', +``European'', +``Middle Eastern'', and + ``Latino (Hispanic)''. As a consequence, one can observe a shift in outward appearance.}
    \label{fig:skin}
\end{figure}

Furthermore, in Fig.~\ref{fig:skin}, we show results for the prompt ``A photo of a woman''. Usually, the generated images by default SD represent people with Caucasian appearance. Here, we generated images with \textsc{Fair Diffusion} by instructing with +``Asian'', +``Indian'', +``African'', +``European'', +``Middle Eastern'', and + ``Latino (Hispanic)''. These instructions are taken from FairFace's ``race'' class. One can observe that the outward appearance changes according to the instruction given. Interestingly, one can observe that the instruction refers to multiple features, like hair color and style, lip color, and shapes of nose, cheek, and chin. This figure illustrates the potential capabilities of \textsc{Fair Diffusion} and should motivate further research in more diverse image generation beyond the presented gender biases.

\begin{figure}[b]
    \centering
    \includegraphics[width=0.65\linewidth]{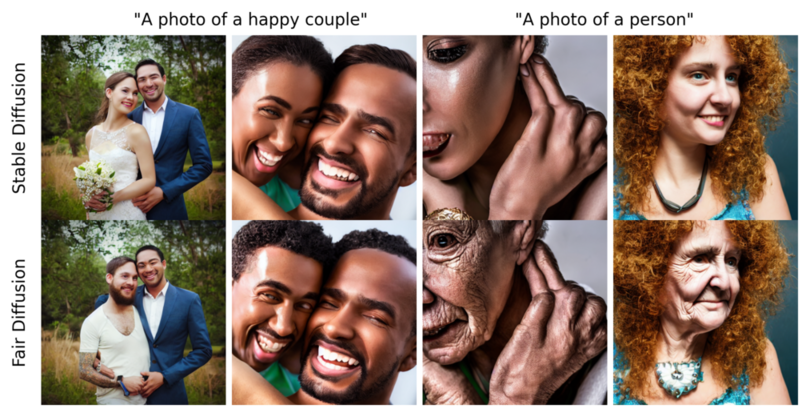}
    \caption{\textsc{Fair Diffusion} approaches heteronormativity or age biases. The top row is generated with default SD and the bottom row with \textsc{Fair Diffusion}. The first two columns are generated with ``A photo of a happy couple'' and instructed with -``heterosexual'' +``homosexual'' and the left two images are generated with ``A photo of a person'' and instructed with -``young person'' +``old person''. As one can observe, the outward appearance of the generated persons changes according to the instructions used.}
    \label{fig:other_biases}
\end{figure}

We found further discriminating behavior in default SD, shown in Fig.~\ref{fig:other_biases}. As one can observe, SD tends to generate images in line with heteronormativity and images of younger persons. Instead, \textsc{Fair Diffusion} can be employed again to generate homosexual couples and people of different ages. Please note that these images are only illustrative to show the potential of \textsc{Fair Diffusion}. As elaborated in the Discussion section, certain dangers also come along. For example, ``homosexual'' and ``gay'' often generated male-homosexual couples. This might be due to the fact that there is a specific word for female homosexuality, i.e., lesbian, and with +``lesbian'', it is also possible to generate female couples. However, these results are preliminary, showing avenues for future research.\footnote{We down-scaled all images to a smaller and easier-to-handle size. Higher-resolution images can be generated with our code.}

\subsection*{B \quad A more detailed Inspection of CLIP Biases}
In this experiment, we tested the similarity between images of different concepts. In line with \cite{steed21image}, we applied their setup to the CLIP encoder and extended their occupation experiments with engineering and caregiving. Therefore, we added the first 12 images from Google search, that did not contain people, to the set of evaluation images of \cite{steed21image}. All images can be found in our code base\footnote{anonymous link} to reproduce the results.

Besides the results shown in Tab.~\ref{tab:iEAT_Clip} we show further results in Tab.~\ref{tab:iEAT_Clip_full}. One can observe further cultural and racial biases for people looking Arab-Muslim or African-American as CLIP relates them to unpleasant. Accordingly, an intersectionality bias \cite{wang22intersectionality} is present in CLIP too, amplifying the gender occupation bias. In other words, in CLIP, some people are confronted with multiple factors of advantage (e.g. White male) or disadvantage (e.g. Black female). Overall, we find that CLIP is inherently affected by bias too, and can be attributed as a source for the bias shift between LAION-5B and SD.

\begin{figure}[t]
    \centering
    \includegraphics[width=0.9\linewidth]{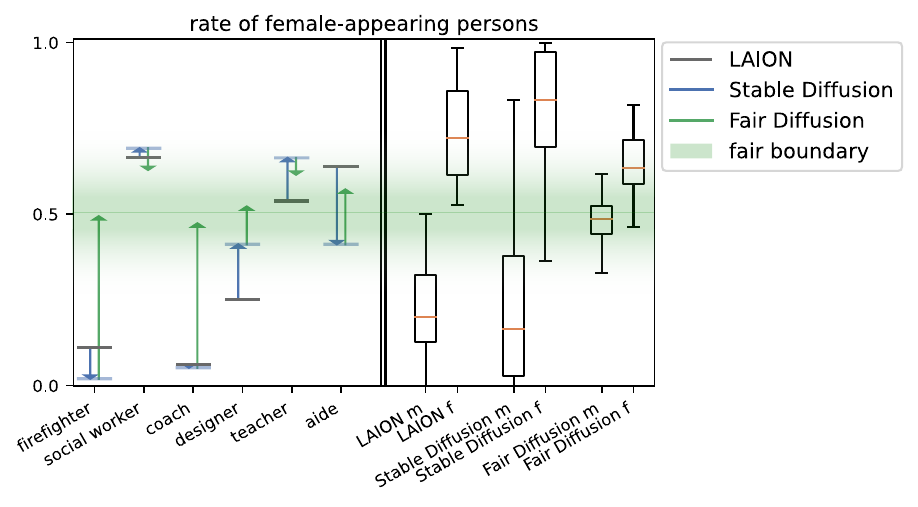}
    \caption{Rate of female-appearing persons for six specific and the per-group median for all 150 occupations.
    Here, \textsc{Fair Diffusion} uses +'male person' and +'female person', respectively, as fair instructions. \textsc{Fair Diffusion} generated images (ours, green arrow) are still closer to the fair boundary than SD and LAION. However, the results are worse than the setup used in Fig.~\ref{fig:gender_proportion}. This shows, that positive guidance alone is already a good way to increase fairness, but negative guidance helps improve performance. (Best viewed in color)}
    \label{fig:gender_proportion2}
\end{figure}

\begin{table}[t]
    \centering
    \caption{Bias Inspection for CLIP. We examine the iEAT for gender occupation biases. The table shows that such biases are present in CLIP, i.e., male-appearing persons are considered to be closer to career, science, or engineering compared to female-appearing who are closer to family, arts, and caregiving. All examples have a high effect size, $d$, along with high significance, i.e., low $p$ value. Furthermore, we evaluated intersectionality biases and found that skin color attributes amplify gender occupation biases. Lastly, broader ethnicity (cultural) biases can be observed for (un)pleasant. $n_t$ and $n_a$ denote the number of images evaluated for each concept.
    }
    \begin{tabular}{c|cccccc}
         topic & target concept & attribute concept & $n_t$ & $n_a$ & p & d \\
         \hline
         Gender & Male - Female & Career - Family & 40 & 21 & 0.01 & 0.58 \\
         Gender & Male - Female & Science - Arts & 40 & 21 & 0.003 & 0.63 \\
         Gender & Male - Female & Engineering - Caregiving & 40 & 12 & 0.005 & 0.57 \\
         Ethnicity & White Male - Black Female & Career - Family & 3 & 21 & 0.1 & 0.99 \\
         Ethnicity & White Male - Black Female & Science - Arts & 3 & 21 & 0.05 & 1.48 \\
         Ethnicity & White Male - Black Female & Engineering - Caregiving & 3 & 12 & 0.05 & 1.57 \\
         Ethnicity & Other people - Arab-Muslim & Pleasant - Unpleasant & 10 & 55 & 0.002 & 1.18 \\
         Ethnicity & European American - African American & Pleasant - Unpleasant & 6 & 55 & 0.002 & 1.66 \\
    \end{tabular}
    \label{tab:iEAT_Clip_full}
\end{table}

\subsection*{C \quad Different Setups for \textsc{Fair Diffusion}}
\label{sec:dif-setups}
\begin{figure}[t]
    \centering
    \includegraphics[width=0.9\linewidth]{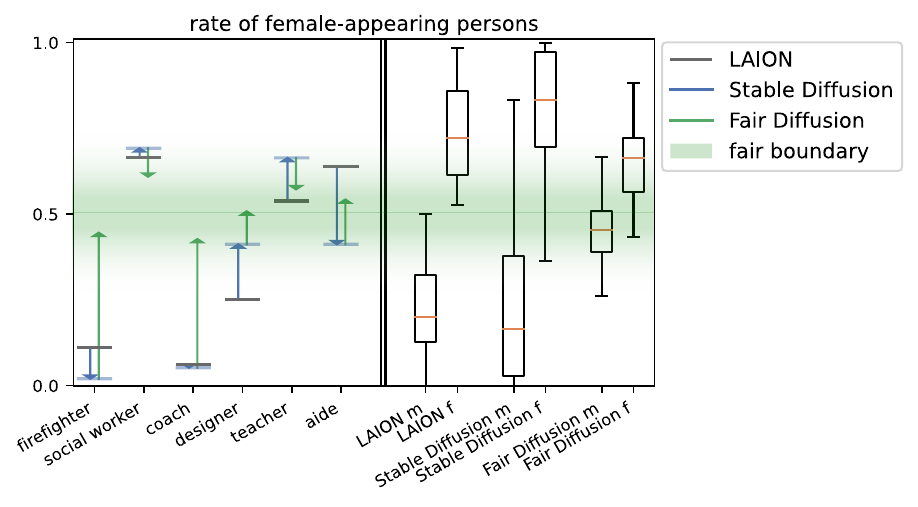}
    \caption{Rate of female-appearing persons for six specific and the per-group median for all 150 occupations.
    Here, \textsc{Fair Diffusion} uses -'male person, female person' +'female person' and -'male person, female person' +'male person', respectively, as fair instructions. \textsc{Fair Diffusion} generated images (ours, green arrow) are still closer to the fair boundary than SD and LAION. However, the results are worse than the setup used in Fig.~\ref{fig:gender_proportion}. This shows that positive and negative guidance interfere with each other, and distinct guidance concepts are required to optimize performance. (Best viewed in color)}
    \label{fig:gender_proportion0}
\end{figure}

\begin{figure}[t]
    \centering
    \includegraphics[width=0.9\linewidth]{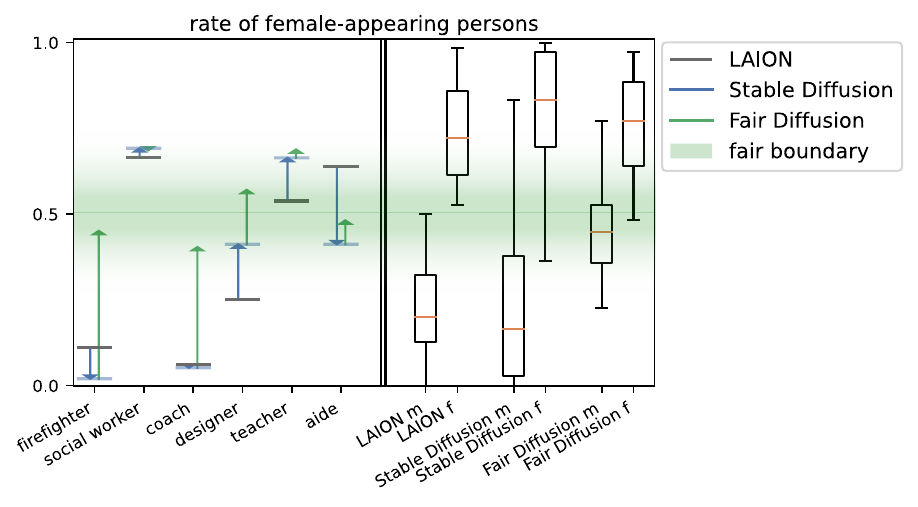}
    \caption{Rate of female-appearing persons for six specific and the per-group median for all 150 occupations.
    Here, \textsc{Fair Diffusion} uses -'gender' +'female gender' and -'gender' +'male gender', respectively, as fair instructions. \textsc{Fair Diffusion} generated images (ours, green arrow) are still closer to the fair boundary than SD, but only partly for LAION. Moreover, the results are worse than the setup used in Fig.~\ref{fig:gender_proportion}. This result suggests that the underlying components of \textsc{Fair Diffusion}, i.e., SD's components, are incapable of the concept 'gender'. Hence, further research is needed here. (Best viewed in color)}
    \label{fig:gender_proportion3}
\end{figure}

\begin{figure}[t]
    \centering
    \includegraphics[width=0.9\linewidth]{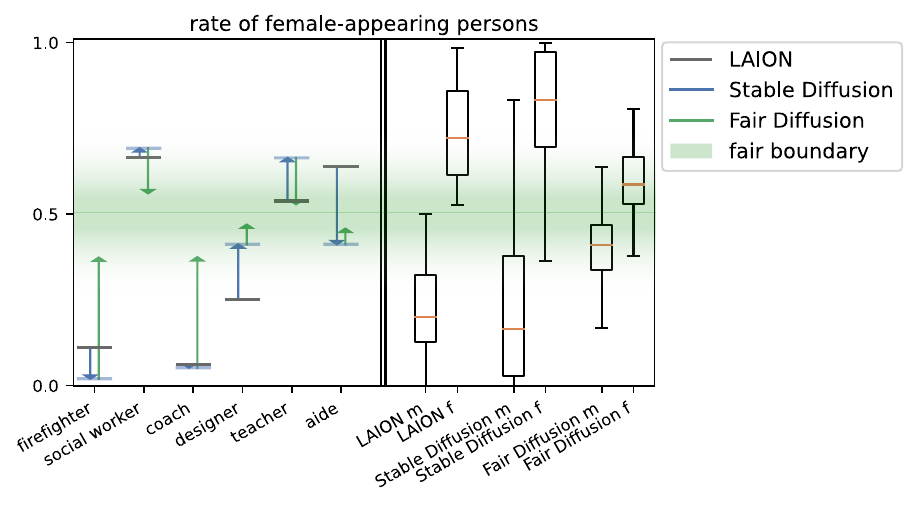}
    \caption{Rate of female-appearing persons for six specific and the per-group median for all 150 occupations.
    Here, \textsc{Fair Diffusion} uses -'female person, male person, non-binary person' +'female person' and -'female person, male person, non-binary person' +'male person', respectively, as fair instructions. \textsc{Fair Diffusion} generated images (ours, green arrow) are still closer to the fair boundary than SD and LAION. However, the results are worse than the setup used in Fig.~\ref{fig:gender_proportion}. This result is also in line with Fig.~\ref{fig:gender_proportion2}, showing that positive and negative guidance interfere with each other and distinct guidance concepts are required to optimize performance. The influence of 'non-binary person' seems small, perhaps due to a lack in SD's understanding of this concept ---further research is needed here. (Best viewed in color)}
    \label{fig:gender_proportion4}
\end{figure}

In addition to the results shown in Fig.~\ref{fig:gender_proportion}, we investigated different setups for \textsc{Fair Diffusion}. The edit instructions used in the main text (-``female person'' +``male person'', and with switched signs) represent a limitation, as examined in Discussion. Hence, we also examined other setups shown in Figs.~\ref{fig:gender_proportion0}, \ref{fig:gender_proportion2}, \ref{fig:gender_proportion3} and \ref{fig:gender_proportion4}. The applied edit instructions are given below each figure. In general, one can observe that the overall performance of \textsc{Fair Diffusion} for the different edit instructions presented here is worse than for the instructions used in the main text. 

However, we cannot clearly attribute the loss in performance to \textsc{Fair Diffusion}. For one, the measured rate of female-appearing persons depends on the FairFace classifier. In case the edit instructions led to less clearly identifiable generated persons, FairFace struggled to classify them correctly and had high uncertainty. Furthermore, the success of the edit instruction depends on the parameters used. We used rather low parameters to not alter the image too strongly. If one increases the parameters, the changes are enforced stronger, at the expense of more substantial changes to the re-generated image. On the other hand, the instructions given to \textsc{Fair Diffusion} have to be known by the model. SD seems to have only a little understanding of e.g. the words ``non-binary'' (Fig.~\ref{fig:gender_proportion4}) and ``gender'' (Fig.~\ref{fig:gender_proportion3}). Thus, it is difficult to appropriately use these concepts for steering the image toward fairer outcomes. Furthermore, Figs.~\ref{fig:gender_proportion2} and \ref{fig:gender_proportion4} show that fair instructions should be distinctive. If they contain similar concepts, they might interfere with each other. Lastly, (Fig.~\ref{fig:gender_proportion0}) shows that positive guidance alone is insufficient compared to positive and negative but still achieves remarkable performance.
More research is needed here to investigate these findings further.

\begin{figure}
    \centering
    \includegraphics[width=0.85\linewidth]{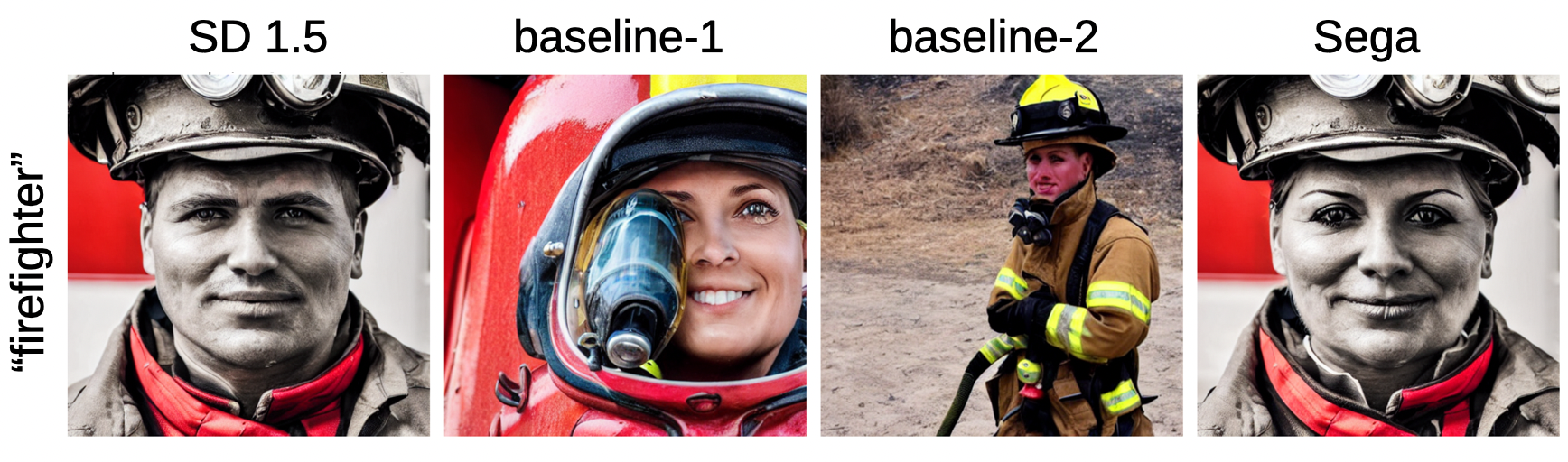}
    \caption{Comparing different image editing techniques. The leftmost image is generated by SD 1.5 with the prompt ``A photo of the face of a firefighter''. To address the present gender occupation bias, we compare three methods: the two most promising methods from Bansal~\etal~\cite{bansal2022diversify}, i.e. baseline-1 (``A photo of the face of a female firefighter'') and baseline-2 (``A photo of the face of a firefighter if all individuals can be a firefighter irrespective of their gender''), and \textsc{Sega} \cite{brack2023Sega}, i.e. semantic guidance (increasing ``female'' and decreasing ``male''). As can be seen, the baselines completely change the image composition and even fail in clearly changing the gender appearance. \textsc{Sega}, in contrast, prevails the image composition while only addressing the concept to be edited.}
    \label{fig:baseline_comparison}
\end{figure}

\paragraph{Comparing Different Image Editing Techniques}
In Fig.~\ref{fig:baseline_comparison}, we compare three different image editing techniques for diffusion models. \textsc{Fair Diffusion} is agnostic to the underlying method and able to integrate various approaches --thus also the three shown. However, the capabilities of the methods differ. One can observe in this exemplary comparison, that only \textsc{Sega} is able to preserve the image composition and edit the gender appearance in isolation. The other methods in contrast change the overall image composition and struggle with addressing the gender appearance.

\subsection*{D \quad Further Experimental Details}
\begin{figure}[t]
    \centering
    \includegraphics[width=0.95\linewidth]{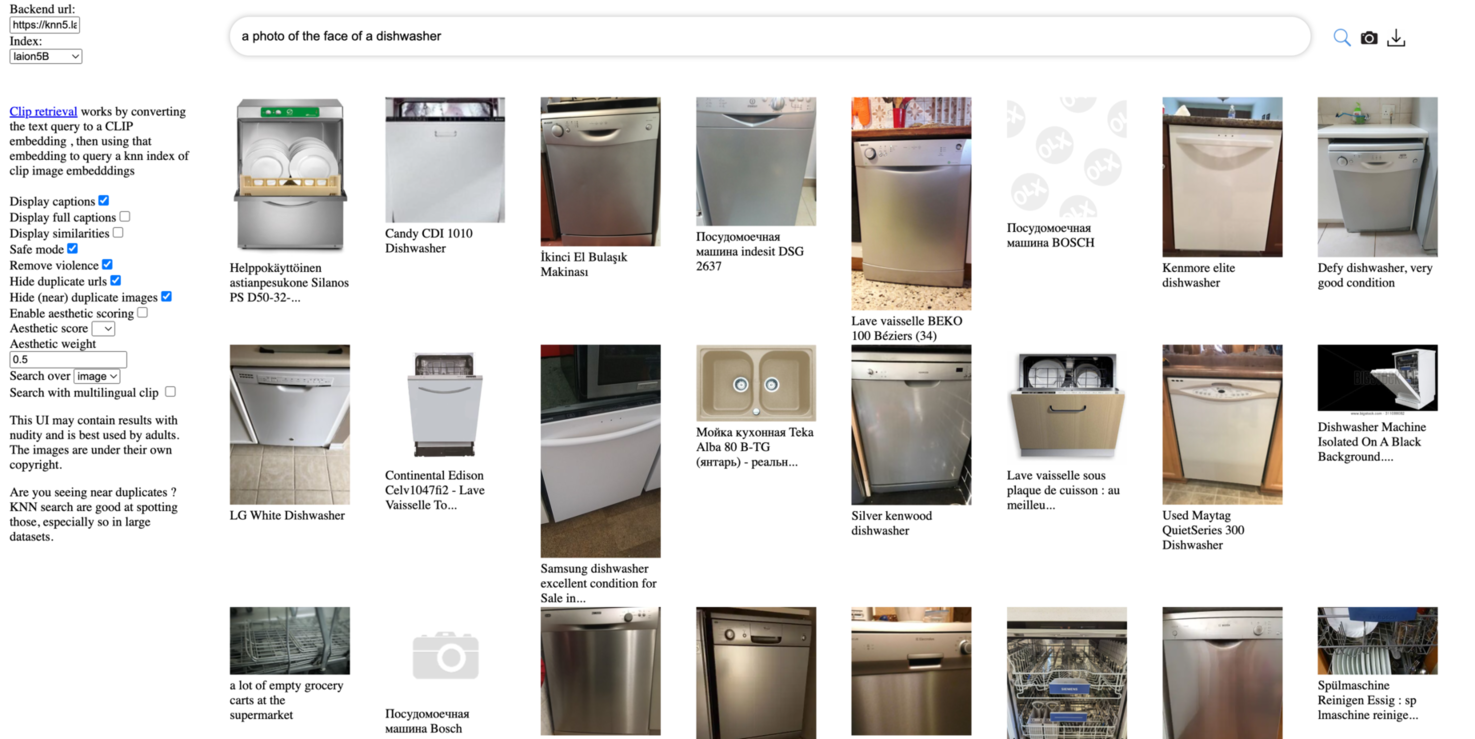}
    \caption{Dishwasher artifact as lack of training data in LAION-5B and the ambiguity of the term ``face of a dishwasher''.}
    \label{fig:dish_artif}
\end{figure}

\paragraph{Artifacts in Image Generation}
As described in our experimental evaluation, we also stumbled upon some challenges. For example, it was difficult to generate images for the occupation ``dishwasher''. In Fig.~\ref{fig:dish_artif}, one can observe that ``face of a dishwasher'' is very ambiguous and mainly yielded results of the front side of dishwashing machines. Hence, further prompts beyond ``A photo of the face of a {$occ$}'' should be evaluated in future research.

\paragraph{Selection of Subgroups}
We hand selected subgroups ``science'', ``arts'', ``engineering'', and ``caregiving''. The selection can be found in Tab.~\ref{tab:occupation_group_list}.

\begin{table}[t]
    \centering
    \caption{Hand-selected subgroups of the whole occupation list. The list is divided into four subgroups: Science, Arts, Engineering, and Caregiving. Each subset contains between 50K and 120K images.}
    \begin{tabular}{cccc}
         Science & Arts & Engineering & Caregiving \\
         \hline
         aerospace engineer & artist & aerospace engineer & childcare worker\\
         claims appraiser & author & architect & coach \\
         clerk & designer & civil engineer & dental assistant\\
         computer programmer & interior designer & computer programmer& dental hygienist\\
         electrical engineer & musician & computer support specialist & dentist\\
         scientist & painter & computer systems analyst & doctor\\
         & photographer & electrical engineer & housekeeper\\
         & singer & engineer & maid\\
         & writer & industrial engineer & massage therapist\\
         & graphic designer & mechanical engineer & mental health counselor\\
         & & programmer & nurse\\
         & & software developer & nursing assistant\\
         & & & occupational therapist \\
         & & & physical therapist\\
         & & & psychologist\\
         & & & social assistant\\
         & & & social worker\\
         & & & teacher\\
         & & & teaching assistant\\
         & & & therapist
    \end{tabular}
    \label{tab:occupation_group_list}
\end{table}

\end{document}